\documentclass[conference]{IEEEtran}
\IEEEoverridecommandlockouts
\usepackage{cite}
\usepackage{amsmath,amssymb,amsfonts}
\usepackage{algorithmic}
\usepackage{graphicx}
\usepackage{textcomp}
\usepackage{xcolor}
\usepackage{booktabs}
\usepackage{pifont}
\usepackage{hyperref}

\newcommand{\cmark}{\ding{51}}
\newcommand{\xmark}{\ding{55}}
\def\BibTeX{{\rm B\kern-.05em{\sc i\kern-.025em b}\kern-.08em
    T\kern-.1667em\lower.7ex\hbox{E}\kern-.125emX}}

\begin{document}

\newcommand{\ModelName}{\textcolor{black}{MMASC}}

\title{Motion-Adaptive Multi-Scale Temporal Modelling with Skeleton-Constrained Spatial Graphs for Efficient 3D Human Pose Estimation\\
}

\author{
\IEEEauthorblockN{
Ruochen Li,
Shuang Chen,
Wenke E,
Farshad Arvin,
Amir Atapour-Abarghouei
}
\IEEEauthorblockA{
Department of Computer Science, Durham University, UK\\
\{ruochen.li, shuang.chen, wenke.e, farshad.arvin, amir.atapour-abarghouei\}@durham.ac.uk
}
}



\maketitle

\begin{abstract}
Accurate 3D human pose estimation from monocular videos requires effective modelling of complex spatial and temporal dependencies. However, existing methods often face challenges in efficiency and adaptability when modelling spatial and temporal dependencies, particularly under dense attention or fixed modelling schemes. In this work, we propose \textbf{MASC-Pose}, a Motion-Adaptive multi-scale temporal modelling framework with Skeleton-Constrained spatial graphs for efficient 3D human pose estimation. Specifically, it introduces an Adaptive Multi-scale Temporal Modelling (AMTM) module to adaptively capture heterogeneous motion dynamics at different temporal scales, together with a Skeleton-constrained Adaptive GCN (SAGCN) for joint-specific spatial interaction modelling. By jointly enabling adaptive temporal reasoning and efficient spatial aggregation, our method achieves strong accuracy with high computational efficiency. Extensive experiments on Human3.6M and MPI-INF-3DHP datasets demonstrate the effectiveness of our approach.
\end{abstract}


\section{Introduction}

3D human pose estimation seeks to recover articulated 3D joint coordinates from monocular 2D images or video sequences, and serves as a fundamental component in a wide range of computer vision applications, including human action recognition \cite{zhang2022ActionHuman,liu2017enhanced,liu2018recognizing}, human-computer interaction \cite{rossol2015multisensor,huo20233d,shotton2011real}, and virtual reality \cite{hagbi2010shape}. Despite its broad applicability, this task remains highly challenging. The complex spatial-temporal dependencies among articulated body joints raise significant challenges, as spatial correlations across limbs and temporal motion patterns are tightly coupled \cite{lange2008role,fredericksen1993spatio}. This coupling makes it difficult to model fine-grained joint interactions and long-range motion dynamics, which is ill-posed and benefits from structured priors that simplify the learning objectives~\cite{abarghouei2019monocular}.

Recent advances in 3D human pose estimation increasingly adopt Transformer-based architectures to model spatial-temporal dependencies among body joints~\cite{zheng2021poseformer,zhao2023poseformerv2,zhang2022mixste,tang2023stcformer,liu2025tcpformer}. Representative works such as PoseFormer~\cite{zheng2021poseformer} and MixSTE~\cite{zhang2022mixste} leverage Transformer architectures to model spatial-temporal dependencies, capturing joint-wise correlations and long-range motion dynamics through self-attention mechanisms. As a result, these approaches achieve strong performance through global motion aggregation, but largely rely on vanilla Transformer designs with limited inductive biases. In parallel, motivated by the structured topology of the human skeleton, graph-based methods model joints as nodes and skeletal connections as edges to encode spatial priors.
Early work such as ST-GCN~\cite{yan2018STGCN} demonstrates the effectiveness of Graph Neural Networks (GNNs) in capturing human skeletal structure, and subsequent studies further integrate GNNs with Transformer architectures to enhance spatial interaction modelling~\cite{peng2024ktpformer,mehraban2024motionagformer,pan20253dGCNFormer}. By explicitly encoding skeletal connectivity, these approaches introduce structural inductive biases that improve spatial reasoning and reduce the reliance on dense attention.

\begin{figure}[t]
\centering
\includegraphics[width=\columnwidth]{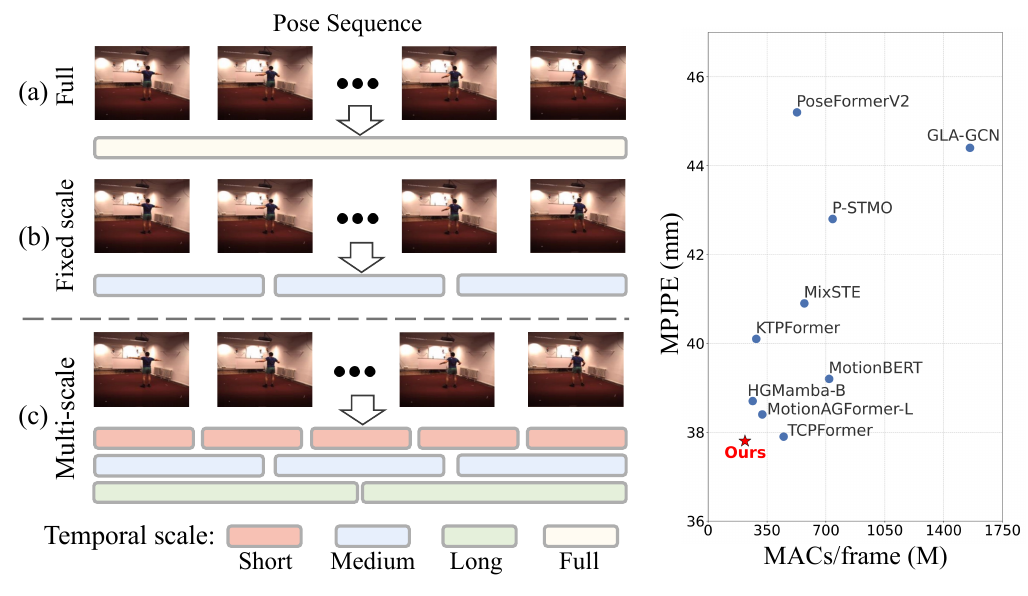} 
\caption{Comparison of temporal dependency modelling strategies and efficiency–accuracy trade-offs.
\textbf{Left}: (a) full-sequence, (b) fixed-scale, and (c) the proposed multi-scale temporal modelling.
\textbf{Right}: Comparison of recent methods on Human3.6M in terms of MPJPE (lower is better) versus MACs/frame, showing that our method achieves state-of-the-art performance with low computational cost.
}
\label{fig:teaser_image}
\end{figure}

Despite their strong performance, existing methods still exhibit notable limitations in spatial-temporal learning. 
For \textbf{temporal dependency modelling}, most approaches rely on global temporal attention \cite{peng2024ktpformer,zheng2021poseformer,zhu2023motionbert,mehraban2024motionagformer} as shown in Figure~\ref{fig:teaser_image} (a). 
While effective in capturing long-range dependencies, applying global temporal attention over long sequences incurs high computational cost and tends to dilute fine-grained temporal saliency due to uniform token aggregation \cite{dong2021attentionnot}. 
TCPFormer~\cite{liu2025tcpformer} partially alleviates this issue by capturing fine-grained temporal patterns through fixed-length temporal abstractions (Figure~\ref{fig:teaser_image}(b)); however, the temporal scale remains fixed and dense attention still incurs quadratic complexity and limited adaptability to diverse motion dynamics. For \textbf{spatial joint interaction modelling}, early attention-based methods apply attention over all joints to capture spatial interactions, but stacking attention layers often leads to over-smoothing \cite{wu2023demystifying}, where joint representations become increasingly similar as depth increases. 
Recent approaches \cite{peng2024ktpformer,mehraban2024motionagformer} integrate graph neural networks with skeleton topology. However, their fixed fusion strategies make it difficult to adapt to joint-specific needs in balancing local feature propagation.

In this paper, we propose \textbf{MASC-Pose} for 3D human pose estimation. For the \textbf{temporal aspect}, MASC-Pose introduces an Adaptive Multi-scale Temporal Modelling (AMTM) module to capture motion dynamics at diverse temporal scales. The key insight is that human motion exhibits heterogeneous temporal patterns, where both short-term dynamics and long-term trends are essential. Accordingly, AMTM processes the input sequence through parallel temporal branches with different scales to model multi-scale motion patterns (Figure~\ref{fig:teaser_image} (c)). Inspired by the mixture-of-experts (MoE) paradigm~\cite{fedus2022switch,li2026vite}, we incorporate a learnable fusion mechanism that dynamically weights each temporal scale based on human pose characteristics, while employing lightweight temporal GCNs within each scale to further reduce computational overhead. For the \textbf{spatial aspect}, we propose a Skeleton-constrained Adaptive GCN (SAGCN) to model joint interactions. SAGCN introduces learnable balancing weights that adaptively regulate the contributions of self-features and neighbour aggregation at each layer for joint-specific spatial interaction modelling. The right panel of Figure~\ref{fig:teaser_image} demonstrates the superior performance and efficiency of our approach. \textit{The source code is available on \href{https://github.com/Carrotsniper/MASC-Pose}{GitHub}.} Our contributions are:
\begin{itemize}
    \item We propose MASC-Pose, a novel spatial-temporal framework for 3D human pose estimation that achieves strong performance with high computational efficiency.
    \item We introduce an Adaptive Multi-scale Temporal Modelling (AMTM) module that enables efficient and expressive modelling of temporal dependencies at multiple temporal scales (Section \ref{sec:AMTM}), which adaptively balances short-term dynamics and long-term motion dynamics.
    \item We propose a Skeleton-constrained Adaptive GCN (SAGCN) for efficient and adaptive spatial joint interaction modelling (Section \ref{sec:SAGCN}), which enables joint-specific feature aggregation.
\end{itemize}

\section{Related Work}
\subsection{Transformer-based Methods}

With the success of Transformer architectures~\cite{vaswani2017attention} in natural language processing, self-attention has become a dominant paradigm for modelling spatial-temporal dependencies in 3D human pose estimation. Early Transformer-based approaches typically adopt factorized spatial-temporal attention, where spatial attention captures joint-wise relationships within each frame and temporal attention models motion dynamics across frames~\cite{zheng2021poseformer,zhao2023poseformerv2,zhang2022mixste,tang2023stcformer}. To enhance representation capacity, MotionBERT~\cite{zhu2023motionbert} leverages large-scale motion pretraining to learn more robust spatial-temporal representations. More recent studies such as TCPFormer~\cite{liu2025tcpformer} further explore temporal abstraction strategies to capture fine-grained short-term motion details and long-range dependencies jointly. Overall, these works demonstrate the effectiveness of Transformer-based architectures for pose estimation. However, in contrast to these generic temporal modelling schemes, MASC-Pose introduces a motion-adaptive multi-scale temporal module together with skeleton-constrained spatial aggregation to achieve efficient and adaptable spatial-temporal representation learning.

\subsection{Graph-based Methods}
Motivated by the structured topology of the human skeleton, graph-based methods represent joints as nodes and skeletal connections as edges to model spatial dependencies. GNN have been widely adopted to capture joint interactions and encode skeletal priors in 3D human pose estimation. Early work such as GLA-GCN \cite{yu2023glagcn} combines GNNs with Temporal Convolutional Networks to capture spatial and temporal correlations. Recent approaches integrate GNNs with Transformer architectures~\cite{peng2024ktpformer,mehraban2024motionagformer,pan20253dGCNFormer,li2025unified}, leveraging graph-based spatial modelling to complement attention-based temporal representations. By explicitly modelling joint connectivity, these methods enhance spatial interaction modelling and reduce the computational overhead associated with dense attention mechanisms. Building on this line of work, MASC-Pose retains the skeletal topology prior but makes the spatial message passing more adaptive, while pairing it with motion-aware multi-scale temporal modelling to better handle diverse dynamics under a more efficient spatial-temporal design.

\begin{figure*}[th]
  \centering
  \includegraphics[width=0.90\linewidth]{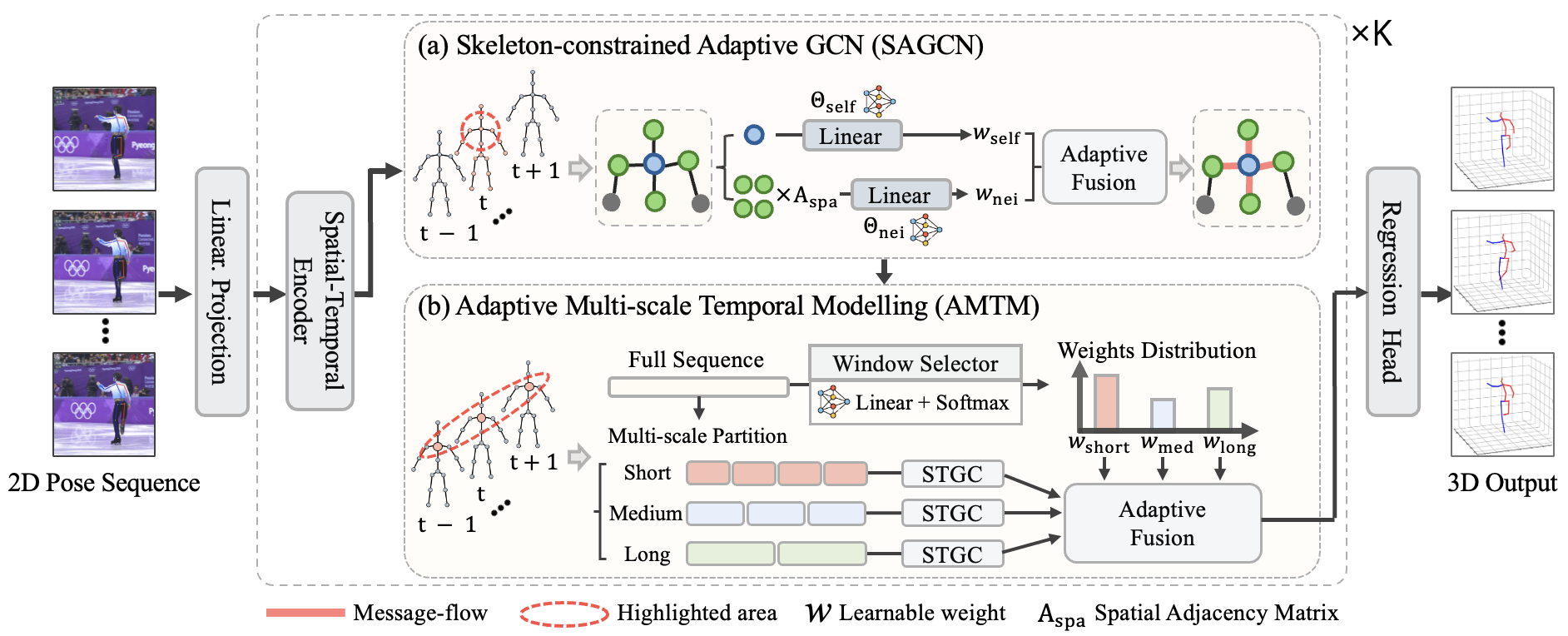}
  \caption{Overview of the proposed framework. The model integrates (a) a Skeleton-constrained Adaptive GCN (SAGCN) for spatial modelling and (b) an Adaptive Multi-scale Temporal Modelling (AMTM) module for temporal modelling. STGC denotes sparse temporal graph convolution operation.}
  \label{fig:overview}
\end{figure*}

\section{Methods}

\subsection{Problem Formulation}
Given an input 2D human pose sequence $\mathbf{X} = \{\mathbf{x}_t\}_{t=1}^{T} \in \mathbb{R}^{T \times J \times 3}$, which contains 2D coordinates and a confidence score of each keypoint. Our objective is to estimate the corresponding 3D pose sequence $\mathbf{Y} = \{\mathbf{y}_t\}_{t=1}^{T} \in \mathbb{R}^{T \times J \times C_{\text{out}}}$. Here, $T$ denotes the number of frames, $J$ denotes the number of body joints per frame, $C_{\text{in}}$ and $C_{\text{out}}$ represent the dimensionality of 2D and 3D joint coordinates, respectively. We first project the input 2D poses into high-dimensional feature space as $\mathbf{X}_{h} \in \mathbb{R}^{T \times J \times D}$ and incorporate positional encoding as $
\mathbf{X}_{st} = \Phi\big( \mathbf{X}_{h} + \mathbf{P}_{\text{pos}} \big)
$, where $\mathbf{P}_{\text{pos}} \in \mathbb{R}^{1 \times J \times D}$ denotes the joint-level positional encoding, $D$ is the hidden dimension, and $\Phi(\cdot)$ represents the backbone spatial-temporal encoder following~\cite{liu2025tcpformer,mehraban2024motionagformer}.

\subsection{Skeleton-constrained Adaptive GCN (SAGCN)}
\label{sec:SAGCN}

Existing methods for spatial joint modelling commonly rely on self-attention or graph-based formulations. Attention-based approaches compute dense joint interactions with high computational cost and limited use of skeletal priors~\cite{liu2025tcpformer,zheng2021poseformer}, while graph-based methods explicitly encode skeleton topology but typically adopt fixed aggregation schemes that treat self and neighbour features uniformly~\cite{mehraban2024motionagformer,peng2024ktpformer}.
Such uniform aggregation overlooks joint-wise heterogeneity in motion patterns. Motivated by this observation, we propose a Skeleton-constrained Adaptive GCN (SAGCN) for spatial interaction modelling (Figure~\ref{fig:overview}(a)), which adaptively balances self and neighbour information.

Given input features $\mathbf{X}_{h} \in \mathbb{R}^{T \times J \times D}$, SAGCN processes each frame independently to model spatial joint interactions by decoupling self-transformation and neighbour aggregation:
\begin{equation}
\mathbf{H} = \sigma\Big(
    w_{\text{self}} \cdot 
    \boldsymbol{\Theta}_{\text{self}}(\mathbf{\mathbf{X}_{h}})
    +\, w_{\text{nei}} \cdot 
    \boldsymbol{\Theta}_{\text{nei}}(\mathbf{A}_{\text{spa}} \mathbf{\mathbf{X}_{h}})
\Big), 
\end{equation}
where $\boldsymbol{\Theta}_{\text{self}}$ and $\boldsymbol{\Theta}_{\text{nei}}$ are independent learnable linear transformations, $\mathbf{A}_{\text{spa}} \in \mathbb{R}^{J \times J}$ is the normalised skeletal adjacency matrix derived from the skeleton topology, $w_{\text{self}}$ and $w_{\text{nei}}$ are learnable balancing weights normalised via softmax to control the contribution of self-features and neighbour aggregation, and $\sigma(\cdot)$ denotes the ReLU activation. Unlike standard GCNs that implicitly mix self and neighbour information with fixed contributions, SAGCN allows the network to learn an optimal balance between local joint features and skeletal context. 

\begin{table*}[t]
\setlength{\tabcolsep}{1.5mm} 
\caption{Quantitative comparisons on the Human3.6M dataset. CE indicates whether the model estimates the center frame only. $T$ denotes the number of input frames. MACs/frames represents the number of multiply-accumulate operations per output frame. MPJPE$^{\dag}$ indicates that ground-truth 2D poses are used as input. The best results are shown in bold and the second-best results are underlined.}
\centering
\fontsize{9}{10}\selectfont{
\begin{tabular}{lcccccccc c}
\toprule
Method & Venue & CE & $T$ & Parameter & MACs & MACs/frames 
& MPJPE $\downarrow$ & P-MPJPE $\downarrow$ & MPJPE$^{\dag}\downarrow$ \\
\midrule

MixSTE~\cite{zhang2022mixste} 
& CVPR'22 & \xmark & 243 & 33.6M & 139.0G & 572M & 40.9 & 32.6 & 21.6 \\

P-STMO~\cite{shan2022pstmo} 
& ECCV'22 & \cmark & 243 & 6.2M & 0.7G & 740M & 42.8 & 34.4 & 29.3 \\

PoseFormerV2~\cite{zhao2023poseformerv2} 
& CVPR'23 & \cmark & 243 & 14.3M & 0.5G & 528M & 45.2 & 35.6 & -- \\

GLA-GCN~\cite{yu2023glagcn} 
& ICCV'23 & \cmark & 243 & 1.3M & 1.5G & 1556M & 44.4 & 34.8 & 21.0 \\

MotionBERT~\cite{zhu2023motionbert} 
& ICCV'23 & \xmark & 243 & 42.3M & 174.8G & 719M & 39.2 & 32.9 & 17.8 \\


KTPFormer~\cite{peng2024ktpformer} 
& CVPR'24 & \xmark & 243 & 33.7M & 69.5G & 286M & 40.1 & 31.9 & 19.0 \\

MotionAGFormer-L~\cite{mehraban2024motionagformer} 
& CVPR'24 & \xmark & 243 & 19.0M & 78.3G & 322M & 38.4 & 32.5 & 17.3 \\

HGMamba-B~\cite{cui2025HGMamba} 
& IJCNN'25 & \xmark & 243 & 14.2M & 64.5G & 265M & 38.7 & 32.9 & \textbf{13.2} \\

H2OT + MotionAGFormer~\cite{li2025h2ot}
& TPAMI'25 & \xmark & 243 & 11.7M & 38.9G & - & 38.5 & - & - \\

TCPFormer~\cite{liu2025tcpformer} 
& AAAI'25 & \xmark & 243 & 35.1M & 109.2G & 449M & \underline{37.9} & \textbf{31.7} & \underline{15.5} \\

\midrule
\textbf{Ours} 
& -- & \xmark & 243 & 13.0M & 53.4G & 219M & \textbf{37.8} & \underline{31.8} & 16.0 \\
\bottomrule
\end{tabular}
}
\label{tab:human36m}
\vspace{-3mm}
\end{table*}

\subsection{Adaptive Multi-scale Temporal Modelling (AMTM)}
\label{sec:AMTM}

For temporal modelling in human pose estimation, most existing methods employ global self-attention \cite{peng2024ktpformer,mehraban2024motionagformer,zheng2021poseformer,zhang2022mixste} or fixed temporal scales \cite{liu2025tcpformer} to capture temporal correlations. However, human motion exhibits inherently multi-scale characteristics, as rapid movements like hand gestures require short-term modelling, while periodic patterns such as walking benefit from long-term context. Processing all frames with a single temporal scale may fail to capture this diversity and introduce unnecessary computational overhead. Inspired by the routing mechanism of MoE~\cite{fedus2022switch}, we propose Adaptive Multi-scale Temporal Modelling (AMTM) (Figure~\ref{fig:overview}(b)) that dynamically routes features across multiple temporal scales, implemented via temporal windows of different lengths.

Specifically, AMTM models temporal dependencies using three parallel partitions with different temporal scales such as short-, medium-, and long-term scales. Given the output $\mathbf{H}$ from SAGCN, AMTM employs a lightweight window selector to adaptively estimate the importance of each window:
\begin{equation}
\mathbf{w} = \text{Softmax}\big( \text{MLP}(\text{GAP}(\mathbf{H})) \big),
\end{equation}
where $\text{GAP}(\cdot)$ denotes global average pooling over temporal dimensions, and 
$\mathbf{w} = [w_{\text{short}}, w_{\text{med}}, w_{\text{long}}]$ represents the predicted importance weights of each scale.

For each temporal scale, we apply Sparse Temporal Graph Convolution (STGC)~\cite{mehraban2024motionagformer}. It first constructs a dynamic temporal graph $\mathbf{A}_{\text{temp}}$ by computing pairwise cosine similarities between all timesteps and retaining only the top-$k$ most similar neighbours for each timestep:
\begin{align}
\mathbf{S} &\in \mathbb{R}^{w \times w}, \quad \mathbf{S}(i,j) = \text{cosine}(\mathbf{h}_i, \mathbf{h}_j), \\
\mathcal{N}_k(t) &= \text{TopK}\big(\mathbf{S}(t,:), k\big), \\
\mathbf{A}_{\text{temp}}(t,\tau) &=
\begin{cases}
1, & \tau \in \mathcal{N}_k(t), \\
0, & \text{otherwise}.
\end{cases}
\end{align}

The sparse adjacency matrix is normalised and used to perform temporal graph convolution:
\begin{equation}
\text{STGC}(\mathbf{A}_{\text{temp}}, \mathbf{H}_w)
= \sigma\Big(
\text{BN}\big(
\tilde{\mathbf{A}}_{\text{temp}} \boldsymbol{\Theta}_{\text{nei}}( \mathbf{H}_w)
+ \boldsymbol{\Theta}_{\text{self}} (\mathbf{H}_w)
\big)
\Big),
\end{equation}
where $\tilde{\mathbf{A}}_{\text{temp}}$ is the normalised adjacency matrix, $\boldsymbol{\Theta}_{\text{nei}}$ and $\boldsymbol{\Theta}_{\text{self}}$ are learnable transformations for neighbour and self features, and $\sigma(\cdot)$ is the ReLU activation. The output of each scale is represented as:
\begin{align}
\mathbf{H}_{\text{short}} &= \text{STGC}_{\text{short}}(\mathbf{A}_{\text{temp}}^{\text{short}}, \text{Partition}_{\text{short}}(\mathbf{H})), \\
\mathbf{H}_{\text{med}} &= \text{STGC}_{\text{med}}(\mathbf{A}_{\text{temp}}^{\text{med}}, \text{Partition}_{\text{med}}(\mathbf{H})), \\
\mathbf{H}_{\text{long}} &= \text{STGC}_{\text{long}}(\mathbf{A}_{\text{temp}}^{\text{long}}, \text{Partition}_{\text{long}}(\mathbf{H})),
\end{align}
where $\text{Partition}_{\text{scale}}(\cdot)$ reshapes the temporal dimension into non-overlapping windows to allow for efficient and focused temporal modelling at each scale.
Finally, we adaptively aggregate the multi-scale outputs:
\begin{equation}
\mathbf{X}_{\text{out}} = w_{\text{short}} \cdot \mathbf{H}_{\text{short}} + w_{\text{med}} \cdot \mathbf{H}_{\text{med}} + w_{\text{long}} \cdot \mathbf{H}_{\text{long}},
\end{equation}
where $\mathbf{X}_{\text{out}}$ represents the aggregated feature representation. Finally, following~\cite{mehraban2024motionagformer,liu2025tcpformer}, a softmax-based adaptive fusion is used to combine $\mathbf{x}_{\text{out}}$ and $\mathbf{x}_{\text{st}}$, producing the final embedding $\mathbf{X}_{\text{final}}$, which is fed into a linear regression head to estimate the final 3D pose sequence. A progressive layer fusion is adopted to aggregate features across layers to preserve multi-level representations and improve gradient flow.

\subsection{Loss Function}
Following prior work \cite{liu2025tcpformer}, our model is trained end to end. The final loss function is defined as:
\begin{equation}
\mathcal{L} = \mathcal{L}_{m} + \lambda_{s} \mathcal{L}_{s} + \lambda_{v} \mathcal{L}_{v} + \lambda_{d} \mathcal{L}_{d},
\end{equation}
where $\mathcal{L}_m$ denotes MPJPE, $\mathcal{L}_s$ represents N-MPJPE, $\mathcal{L}_v$ is the velocity loss, and $\mathcal{L}_d$ represents temporal consistency loss. We set $\lambda_s$ to 0.5, $\lambda_v$ to 20, and $\lambda_d$ to 0.5, respectively.

\section{Experiments}
\subsection{Datasets and Evaluation Metrics}
We evaluate our model on two widely used 3D human pose estimation benchmarks, Human3.6M~\cite{h36m} and MPI-INF-3DHP~\cite{3dhp}.
Human3.6M is a standard indoor dataset with multi-view recordings of multiple subjects performing common actions, while MPI-INF-3DHP covers both indoor and outdoor scenes with diverse activities. Following standard evaluation protocols~\cite{liu2025tcpformer,mehraban2024motionagformer}, we report MPJPE and Procrustes-aligned MPJPE (P-MPJPE) on Human3.6M. For MPI-INF-3DHP, ground-truth 2D poses are used as input, and performance is evaluated using MPJPE, Percentage of Correct Keypoints (PCK), and Area Under the Curve (AUC).

\subsection{Implementation Details}
Our model is implemented in PyTorch and trained on a single NVIDIA H200 GPU.
We use 16 encoder layers with a feature dimension of 128.
The input sequence length is 243 for Human3.6M and 81 for MPI-INF-3DHP, with temporal scales set to $\{9,27,81\}$ and $\{3,9,27\}$, respectively.
Data preprocessing and evaluation protocols follow~\cite{liu2025tcpformer}.
Training is performed using AdamW with a learning rate of $5\times10^{-4}$ for 80 epochs and a batch size of 6.

\begin{table}[t]
\centering
\caption{Quantitative comparison on MPI-INF-3DHP dataset. $T$ denotes the number of input frames. 
PCK and AUC are reported in percentage (\%). The best results are shown in bold and the second-best results are underlined.}
\label{tab:mpi3dhp}
\setlength{\tabcolsep}{2.3pt}
\renewcommand{\arraystretch}{1.1}
\begin{tabular}{l l c c c c}
\toprule
Methods & Venue & $T$ & PCK$\uparrow$ & AUC$\uparrow$ & MPJPE$\downarrow$ \\
\midrule
MixSTE~\cite{zhang2022mixste}        & CVPR'22 & 27  & 94.4 & 66.5 & 54.9 \\
P-STMO~\cite{shan2022pstmo}          & ECCV'22 & 81  & 97.9 & 75.8 & 32.2 \\
PoseFormerV2~\cite{zhao2023poseformerv2} & CVPR'23 & 81  & 97.9 & 78.8 & 27.8 \\
GLA-GCN~\cite{yu2023glagcn}           & ICCV'23 & 81  & 98.5 & 79.1 & 27.7 \\
KTPFormer~\cite{peng2024ktpformer}    & CVPR'24 & 81  & 98.9 & 85.9 & 16.7 \\
MotionAGFormer-L~\cite{mehraban2024motionagformer}  & WACV'24 & 81  & 98.2 & 85.3 & 16.2 \\
HGMamba-B~\cite{cui2025HGMamba}       & IJCNN'25 & 81  & 98.7 & \underline{87.9} & \textbf{14.3} \\
H2OT + MotionAGFormer \cite{li2025h2ot} & TPAMI'25 & 81 &\textbf{99.1} &85.2 &18.0 \\
TCPFormer~\cite{liu2025tcpformer}     & AAAI'25 & 81  & \underline{99.0} & 87.7 & \underline{15.0} \\
\midrule
Ours                                 & --      & 81  & \textbf{99.1} & \textbf{88.2} & 15.5 \\
\bottomrule
\end{tabular}

\end{table}

\subsection{Quantitative Comparison with State-of-the-art Methods}
\subsubsection{Results on Human3.6M dataset} 
Table~\ref{tab:human36m} reports comparisons with state-of-the-art methods on the Human3.6M dataset, evaluating both computational cost and prediction accuracy in terms of MPJPE, P-MPJPE and MPJPE$^{\dag}$. Our method achieves the best MPJPE among all compared approaches, while maintaining competitive performance on P-MPJPE and MPJPE$^{\dag}$. Notably, compared with recent transformer-based methods such as KTPFormer and TCPFormer, our approach attains comparable or superior accuracy with substantially fewer parameters and lower computational cost.

\subsubsection{Results on MPI-INF-3DHP dataset} 
Table~\ref{tab:mpi3dhp} reports quantitative comparisons with state-of-the-art methods on the MPI-INF-3DHP dataset using PCK, AUC, and MPJPE. Our method achieves competitive performance across all metrics, attaining the highest AUC and matching the best PCK score with a low MPJPE. Compared with TCPFormer, our approach delivers comparable or improved accuracy with a more efficient spatial-temporal modelling strategy, indicating strong generalization ability to challenging scenarios.

\subsection{Ablation Study and Analysis}
We conduct a series of ablation study on Human3.6M \cite{h36m} to evaluate the effectiveness of main modules of our method.

\subsubsection{Impact of main component}
\begin{table}[t]
\centering
\caption{Ablation study of different model variants on Human3.6M.}
\label{tab:ablation_variant}
\begin{tabular}{l c}
\toprule
Variant & MPJPE $\downarrow$ \\
\midrule
Baseline (w/o AMTM \& SAGCN) & 41.1 \\
Only AMTM & 38.5 \\
Only SAGCN & 39.9 \\
AMTM + SAGCN (w/o adaptive aggregation) & \underline{38.3} \\
AMTM (w/o multi-scale fusion) + SAGCN & 38.6 \\
\midrule
AMTM + SAGCN (\textbf{Ours}) & \textbf{37.8} \\
\bottomrule
\end{tabular}
\end{table}

Table~\ref{tab:ablation_variant} examines the contribution of individual components in our framework on Human3.6M. Relative to the baseline, incorporating AMTM alone leads to a substantial reduction in MPJPE to 38.5, which is notably larger than the improvement achieved by using SAGCN alone, suggesting that temporal modelling plays a more prominent role in this task. Combining both components yields further performance gains, while removing the adaptive mechanisms consistently results in degraded accuracy. specifically, replacing the adaptive multi-scale aggregation operation in AMTM with fixed weights (i.e. $1/3$ for each temporal scale) leads to a performance degradation to 38.6, indicating the importance of adaptive scale weighting mechanism. The full model achieves the best overall performance, demonstrating the complementary contributions of AMTM and SAGCN.

\subsubsection{Effect of Temporal Scale Configuration}

\begin{table}[t]
\centering
\begin{minipage}{0.48\columnwidth}
\centering
\caption{Ablation on temporal scale on Human3.6M with $T=243$.}
\label{tab:temporal_scales}
\begin{tabular}{l c}
\toprule
Temporal scales & MPJPE $\downarrow$ \\
\midrule
$[3,\,9,\,27]$ & 38.7 \\
$[27,\,81,\,243]$ & \underline{38.6} \\
\midrule
$[9,\,27,\,81]$ (\textbf{Ours}) & \textbf{37.8} \\
\bottomrule
\end{tabular}
\end{minipage}
\hfill
\begin{minipage}{0.48\columnwidth}
\centering
\caption{Ablation on SAGCN hop number.}
\label{tab:sagcn_hop}
\begin{tabular}{c c}
\toprule
Hop number ($K$) & MPJPE $\downarrow$ \\
\midrule
$K=4$ & 39.0 \\
$K=3$ & 38.5 \\
$K=2$ & \underline{38.2} \\
\midrule
$K=1$ (\textbf{Ours}) & \textbf{37.8} \\
\bottomrule
\end{tabular}
\end{minipage}
\end{table}

Table~\ref{tab:temporal_scales} presents an ablation study on temporal scale configurations on the Human3.6M dataset with $T=243$. Using small temporal scales $[3,9,27]$ limits the model's ability to capture long-term motion dependencies, while overly larger scales $[27,81,243]$ tend to dilute fine-grained temporal dynamics. The proposed configuration achieves the best performance, indicating that a balanced combination and design of different scales can be crucial for improved performance.

\subsubsection{Impact of SAGCN Hop Configuration}

As shown in Table~\ref{tab:sagcn_hop}, we study the effect of the hop number $K$ in SAGCN, which controls the spatial propagation range, with $K=1$ aggregating information from each joint and its directly connected neighbours defined by the skeleton topology. The results suggest that larger spatial receptive fields do not necessarily improve performance, likely due to over-smoothing effects \cite{wu2023demystifying} caused by increasing the number of GNN layers.

\begin{figure}[t]
\centering
\includegraphics[width=\columnwidth]{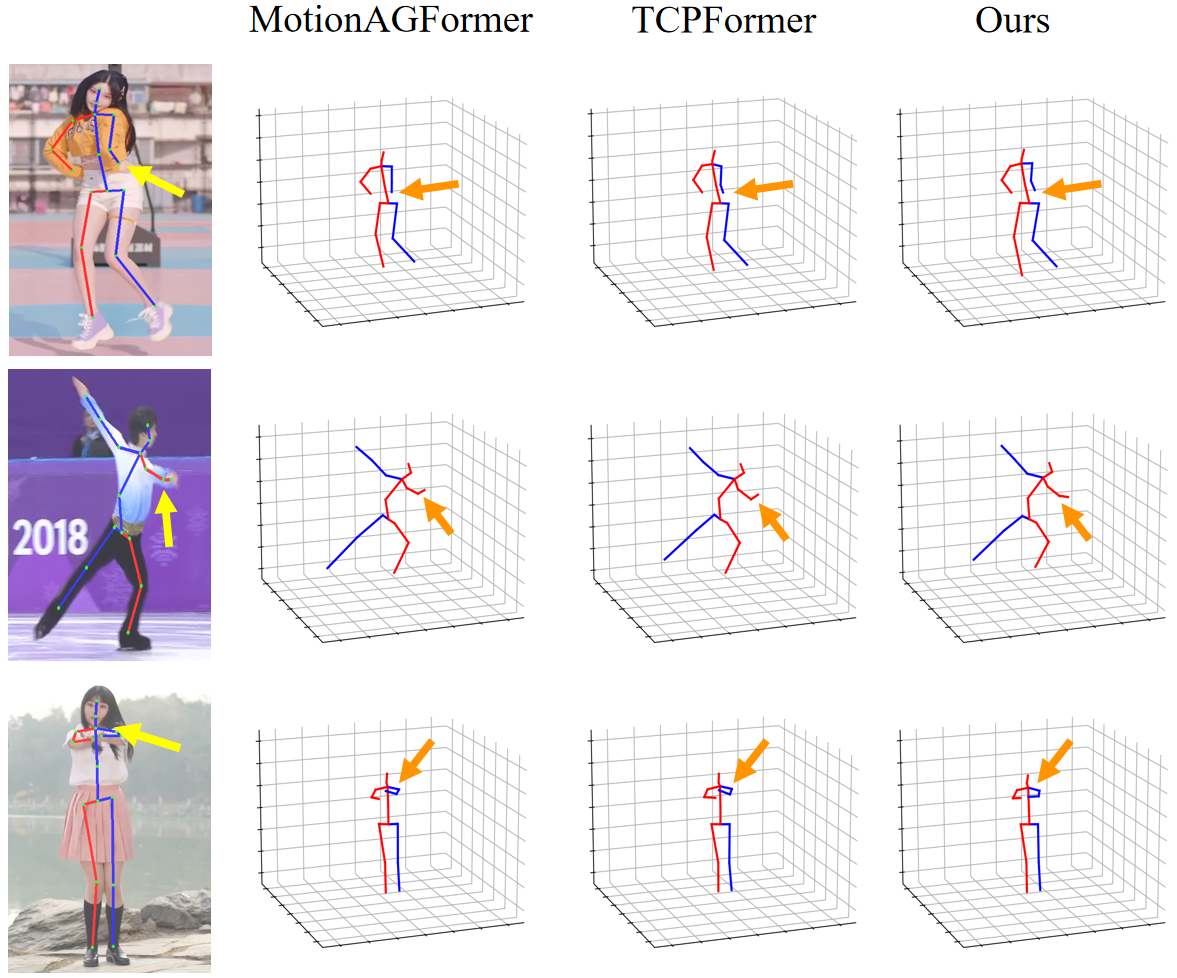} 
\caption{Qualitative comparisons of our method with MotionAGFormer and TCPFormer on in-the-wild videos. We highlight inaccurate or ambiguous 2D detections with light-yellow arrows and indicate the corresponding deviations in the reconstructed 3D poses using orange arrows. }
\label{fig:human36_wild}
\end{figure}

\begin{figure}[t]
\centering
\includegraphics[width=\columnwidth]{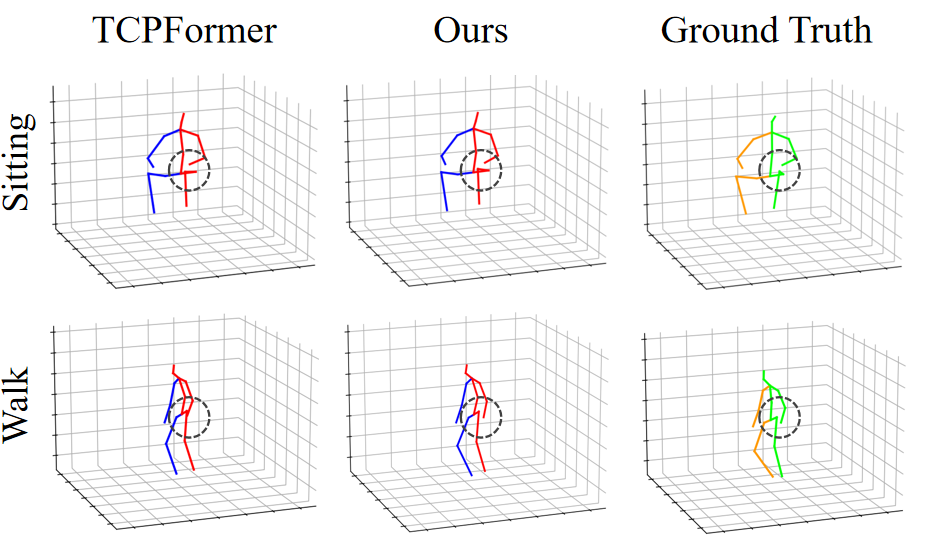} 
\caption{Qualitative comparisons between our method and TCPFormer on the Human3.6M dataset for the \emph{Sitting} and \emph{Walk} actions. Black dashed circles indicate highlighted regions.}
\label{fig:human36}
\end{figure}

\subsection{Qualitative Analysis}

Figure~\ref{fig:human36_wild} shows qualitative comparisons with MotionAGFormer \cite{mehraban2024motionagformer} and TCPFormer \cite{liu2025tcpformer} on in-the-wild videos. Compared methods exhibit noticeable 3D pose deviations under inaccurate or ambiguous 2D detections, whereas our method produces more stable and anatomically plausible pose estimation results. Figure~\ref{fig:human36} shows comparisons between our method and TCPFormer on the Human3.6M dataset for the \emph{Sitting} and \emph{Walk} actions. Our method shows 3D pose estimations that are closer to the ground truth compared to TCPFormer.

\begin{figure}[t]
\centering
\includegraphics[width=0.9\columnwidth]{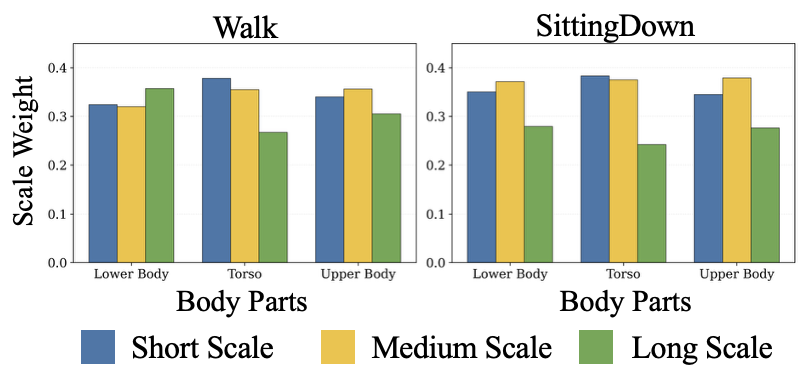} 
\caption{Visualisation of the scale weight distribution for the \emph{Walk} and \emph{SittingDown} actions on the Human3.6M dataset (243 frames). 
Body joints are grouped into \textbf{lower body} (hip, knee, ankle), \textbf{torso} (root, spine, thorax, neck, head), and \textbf{upper body} (shoulder, elbow, wrist). 
Bars represent the average selection weights assigned to three temporal scales: short (9 frames), medium (27 frames), and long (81 frames).}
\label{fig:window_weight}
\end{figure}

As shown in Figure~\ref{fig:window_weight}, we analyse the learned temporal scale weights across different body parts for \emph{Walk} and \emph{SittingDown} actions on Human3.6M with $T=243$. The two actions exhibit distinct temporal characteristics: \emph{Walk} involves periodic and repetitive motion patterns, whereas \emph{SittingDown} is characterised by more abrupt and transitional dynamics. Accordingly, for \emph{Walk}, the lower body assigns higher weights to long temporal scales to capture periodic gait motions, while the torso emphasises short scales for rapid upper-body dynamics. In contrast, for \emph{SittingDown}, all body parts favour short and medium temporal scales, highlighting the importance of intermediate temporal context during posture transitions. These visualisations demonstrate that our model effectively adapts to diverse motion scenarios and enhances the interpretability of the temporal representations.

\section{Conclusion}
We propose MASC-Pose, an efficient spatial-temporal framework for 3D human pose estimation that addresses the limitations of global or fixed-scale temporal modelling while preserving strong accuracy. By adaptively learning both multi-scale temporal correlations and skeleton-constrained spatial interactions, the proposed method achieves a favorable balance between estimation accuracy and computational efficiency. Extensive experiments demonstrate that our method achieves strong performance and robustness under diverse motion patterns. Future work will explore adaptive temporal partitioning with overlapping or variable-length scales for modelling non-stationary motions, as well as broader generalisation settings such as cross-dataset transfer and in-the-wild evaluation.

\bibliographystyle{IEEEtran}
\bibliography{egbib.bib}

\end{document}